%% file: main.tex
\documentclass[10pt,twocolumn,letterpaper]{article}

\usepackage[pagenumbers]{cvpr} 

\input{preamble}

\definecolor{cvprblue}{rgb}{0.21,0.49,0.74}
\usepackage[pagebackref,breaklinks,colorlinks,allcolors=cvprblue]{hyperref}

\usepackage{amsthm}
\usepackage{algorithm}
\usepackage{algpseudocode}
\usepackage{multirow}
\usepackage{colortbl}
\usepackage{capt-of}
\definecolor{pqrsblue}{RGB}{230,240,250}

\def\confName{CVPR}
\def\confYear{2027}

\usepackage{pifont}
\providecommand{\cmark}{\ding{51}}
\providecommand{\xmark}{\ding{55}}

\title{Every Query Routes Itself: Accelerating Video Diffusion Transformers with Per-Query Routed Sparse Attention}

\author{%
  Zhen Huang$^{1,*}$\quad
  Shuwei Li$^{1,*}$\quad
  Ruizhe Yao$^{2}$\quad
  Xinrui Chen$^{1}$\quad
  Danyi Liu$^{1}$\quad
  Haoyuan Liang$^{1}$\\
  Zhiyu Ye$^{1}$\quad
  Qiuyan Sun$^{1}$\quad
  Bingwei Chen$^{3}$\quad
  Yiyuan Li$^{3}$\quad
  Xu Zhou$^{3,\dagger}$\quad
  Haohuan Fu$^{1,\dagger}$\\
  {\small
    $^{1}$Tsinghua University\qquad
    $^{2}$Harbin Institute of Technology, Shenzhen\qquad
    $^{3}$Sangfor Technologies%
  }\\
  {\small\ttfamily
    z-huang26@mails.tsinghua.edu.cn\qquad
    haohuan@tsinghua.edu.cn%
  }\\
  {\small
    $^{*}$Equal contribution.\qquad
    $^{\dagger}$Corresponding authors.%
  }%
}

\begin{document}
\twocolumn[{%
\renewcommand\twocolumn[1][]{#1}%
\maketitle
\vspace{-2em}
\begin{center}
\includegraphics[width=\textwidth]{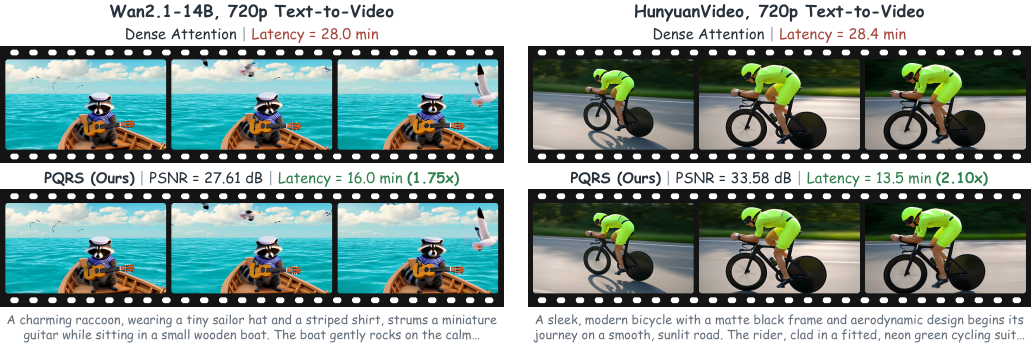}
\captionof{figure}{\textbf{PQRS accelerates 720p video generation while preserving visual quality.} Each block shows frames from dense attention (top) and PQRS (bottom) generated with the same prompt and seed. On a single H100, PQRS reaches 1.75$\times$ and 2.10$\times$ end-to-end speedup on Wan2.1 and HunyuanVideo, with PSNRs of 27.61 and 33.58~dB on the shown samples. Prompts are truncated.}
\label{fig:teaser}
\end{center}%
}]

\input{sec/0_abstract}    
\input{sec/1_intro}
\input{sec/2_related}
\input{sec/3_motivation}
\input{sec/4_method}

\input{sec/5_exp}
\input{sec/6_conclusion}

{
    \small
    \bibliographystyle{ieeenat_fullname}
    \bibliography{main}
}

\input{sec/X_suppl}

\end{document}

%% file: preamble.tex
\usepackage{microtype}



%% file: sec/0_abstract.tex
\begin{abstract}
Self-attention over long token sequences dominates the inference cost of video diffusion transformers. Clustering-based sparse attention clusters both queries and keys and lets all queries in a cluster share one selection of key clusters. We show that this sharing limits accuracy: with key clusters and density fixed, attention grows more accurate as the query clusters are refined, and is most accurate when each query selects on its own, which needs no query clustering at all. We therefore propose \textbf{P}er-\textbf{Q}uery \textbf{R}outed \textbf{S}parse Attention (PQRS), a training-free method that clusters only keys, lets each query select its own key clusters, and compensates for the unselected ones with their centroids. Per-query selections leave no query tile with a shared key list, so PQRS traverses key clusters instead of queries, buckets queries with identical selections into dense tasks, and merges their partial results in one pass under a fixed softmax offset carried from the previous denoising step. On four text-to-video and image-to-video models from the Wan and HunyuanVideo families, PQRS is the fastest of all the sparse attention methods we evaluate and the closest to the dense output, reaching $28.9$ dB PSNR on Wan2.1-T2V and $35.4$ dB on HunyuanVideo at $1.66$--$2.10\times$ end-to-end speedup.
\end{abstract}

%% file: sec/1_intro.tex
\section{Introduction}
\label{sec:intro}
Video diffusion transformers denoise token sequences whose length sets their cost: Wan2.1 \cite{wan2025wan} at 720p with 81 frames denoises $N = 75{,}600$ tokens, and self-attention over them is quadratic in $N$ and accounts for most of the 28 minutes a five-second clip takes on an H100 \cite{yang2025sparse, tan2026adacluster}. Sparse attention reduces this cost by computing only the token pairs that carry most of the attention mass. Most methods select over blocks of contiguous tokens, and contiguity in a video latent does not imply similarity: tokens that respond to the same keys are often far apart in raster order. Clustering-based methods \cite{yang2025sparse, luo2026attention, tan2026adacluster, zhou2026svgear} remove this mismatch by grouping tokens by content, and among training-free methods they currently give the best fidelity at a given cost.

Keys and queries are clustered for different reasons. Keys are clustered so that a query can select at cluster granularity. Queries are clustered so that the selection can be executed: when a whole query cluster shares one selection, each selected pair of clusters becomes a dense block for a block-sparse kernel. But the shared selection also fixes what each query may see, and a query that attends differently from its cluster mean is handed the mean's keys. Holding the key clusters and the density fixed, we find that recall rises and compensated output error falls as the query clusters are refined, with one query per cluster the most accurate. Even when each query cluster is scored by the attention mass it actually receives, so that the score carries no estimation error, a shared selection still trails per-query routing. The sharing itself is what costs accuracy.

Per-query routing needs no query clusters at all, so it removes a step rather than adding one. Query clustering is not free either: a $k$-means pass compares every token against every centroid, the same order of cost as scoring every query against the key centroids once. The finest granularity is therefore not the more expensive option. What stands in the way is the shape of the work rather than its volume: tensor cores multiply a tile of queries against a tile of keys, and such a tile pays off only when its rows attend to the same keys, which is why two-sided clustering groups queries before they select \cite{yang2025sparse}.

We therefore propose \textbf{Per-Query Routed Sparse Attention (PQRS)}, which clusters only the keys and lets every query select on its own. Each query scores the key centroids, ranks the clusters by estimated attention mass, and keeps the top clusters whose cumulative mass reaches $p$; the selected clusters are computed densely, and the rest are compensated through their centroid summaries using the scores the routing has already produced. Since no tile of consecutive queries now shares a key list, our kernel \emph{traverses key clusters} instead: within a group of similar clusters it buckets the queries by which members they chose, so that queries with the same selection form a dense task, and execution is shared only where the hardware requires it. Partial results accumulate in place under a fixed offset taken from the log-sum-exp of the previous denoising step, which removes the rescaling of online softmax and lets the kernel finish in a single pass.

We evaluate PQRS on four 720p video DiTs covering text-to-video and image-to-video generation from the Wan and HunyuanVideo families, with one configuration shared by all models and no training or per-model calibration. Against five training-free sparse attention baselines, PQRS attains the best PSNR, SSIM and LPIPS on every model while also being the fastest, with end-to-end speedups of $1.66\times$ to $2.10\times$ on a single H100. On VBench, which judges a video without reference to the dense output, we detect no degradation relative to dense attention. \Cref{fig:teaser} compares frames from dense attention and PQRS. Our contributions are the following.

\begin{itemize}
\item We locate the accuracy cost of clustering-based sparse attention in the shared selection rather than in the clustering.
\item We propose PQRS, a training-free method in which every query routes over key clusters on its own and compensates the rest through their centroids, together with a cluster-major kernel that executes these irregular selections at tile efficiency in a single pass.
\item On four evaluated 720p video DiTs, PQRS attains the best PSNR, SSIM and LPIPS against five training-free baselines while also being the fastest, with end-to-end speedups of $1.66\times$ to $2.10\times$ and no degradation detected on VBench.
\end{itemize}

%% file: sec/2_related.tex
\section{Related Work}
\label{sec:related}

\paragraph{Sparse attention for video DiTs.}
Static methods exploit spatiotemporal locality in video DiTs~\cite{kong2025hunyuanvideo,wan2025wan} with head-wise spatial or temporal masks~\cite{xi2025sparse}, sliding tiles~\cite{zhang2025fast} or energy-decay masks~\cite{li2025radial}, while dynamic methods estimate block importance online~\cite{zhang2025spargeattention,xu2025xattention} or search the mask at a few steps and reuse it~\cite{xia2025adaspa}. Following content-based clustering in language models~\cite{roy2020efficient,vyas2020clusteredattention}, clustering-based methods permute similar tokens into contiguous blocks: SVG2~\cite{yang2025sparse} clusters queries and keys separately and selects cluster pairs by centroid scores, and SVOO~\cite{luo2026attention}, AdaCluster~\cite{tan2026adacluster}, HyperVAttention~\cite{lee2026hypervattention} and SCOPE~\cite{zhao2026scope} refine its clustering. SVG-EAR~\cite{zhou2026svgear}, PISA~\cite{li2026pisa} and Sol-Attn~\cite{li2026sol} approximate rather than drop unselected blocks, and trainable methods learn the routing~\cite{zhang2025faster,wu2026vmoba}. To use block-sparse kernels, most training-free methods share one key selection across the queries of a block or cluster; TRA~\cite{chen2026tra} assigns per-query budgets but still executes a block-sparse mask, and SparsePR~\cite{taghavi2026sparsepr} regroups queries by their responses to key clusters. PQRS instead removes query grouping: each query routes over shared key clusters on its own.

\paragraph{Query-level sparse attention and its kernels.}
Per-query selection is common in LLMs, both in single-query decoding over KV pages or clusters~\cite{tang2024quest,liu2025clusterkv} and in trainable architectures that route each token to key blocks or tokens~\cite{lu2025moba,yuan2025nsa,deepseekai2025deepseekv32,lai2026msa}; Squeezed Attention~\cite{hooper2025squeezed} also scores key-cluster centroids per query, but averages the scores over a query block in prefill. To execute per-query selections, MoBA~\cite{lu2025moba} reorders queries by key block and merges partial outputs with online softmax, while FSA~\cite{yan2026fsa} and MSA~\cite{lai2026msa} loop over key blocks outermost and merge buffered partial outputs in a separate kernel. PQRS follows this key-outer order but accumulates partial outputs in place with atomic additions, replacing online-softmax rescaling with a fixed per-row exponent offset. The offset is the log-sum-exp of the previous denoising step, which AdaSpa~\cite{xia2025adaspa} observed to be stable across steps, analogous to the forward log-sum-exp reused in the FlashAttention-2 backward pass~\cite{dao2024flashattention} and the shared maximum of FlashDecoding++~\cite{hong2024flashdecoding}. Within a group of similar key clusters, queries with identical selections share one tile rather than the union of their selections, much as Hydragen~\cite{juravsky2024hydragen} batches queries over a shared prefix.

\paragraph{Other acceleration of diffusion models.}
Orthogonal to attention sparsity, fast solvers~\cite{song2021denoising,lu2022dpmsolver} and distillation~\cite{yin2024dmd} reduce the number of sampling steps, feature caching reuses computation across steps~\cite{zhao2025realtime,liu2024timestep}, and quantization lowers numerical precision~\cite{li2025svdquant,zhang2025sageattention}.

%% file: sec/3_motivation.tex
\section{Motivation}
\label{sec:analysis}

\begin{figure}[t]
\centering
\includegraphics[width=\columnwidth]{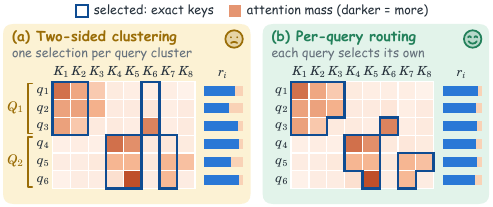}
\caption{\textbf{Shared selection versus per-query routing.} Darker cells carry more attention mass, outlined cells are selected, and bars show the recall $r_i$. (a) A shared selection fits some queries in a cluster poorly. (b) Per-query routing keeps a similar recall for every query with one fewer selected pair.}
\label{fig:shared-vs-perquery}
\end{figure}

Clustering-based sparse attention makes all queries in a cluster share one selection of key clusters, whereas \emph{per-query routing} lets every query select its own, as \Cref{fig:shared-vs-perquery} illustrates. After \Cref{sec:prelim} formalizes the shared selection, \Cref{sec:accuracy} shows that accuracy improves as the query clusters are refined down to single queries, which yields per-query routing. \Cref{sec:cost} shows that per-query routing also avoids clustering the queries, but that its irregular selections do not map onto the tiles of GPU attention kernels.

\subsection{Preliminaries: Clustered Sparse Attention}
\label{sec:prelim}

\paragraph{Sparse attention.} Consider one attention head over $N$ tokens, whose queries, keys, and values $Q, K, V \in \mathbb{R}^{N \times D}$ have rows $q_i$, $k_j$, and $v_j$. Let $s_{ij} = q_i k_j^\top / \sqrt{D}$ be the attention logit and $P_{ij} = e^{s_{ij}} / \sum_{j'} e^{s_{ij'}}$ the dense attention probability, so the dense output is $o_i = \sum_j P_{ij} v_j$. Sparse attention restricts the softmax to the pairs selected by a mask $M \in \{0,1\}^{N \times N}$. Its cost scales with the density $\rho = \frac{1}{N^2} \sum_{i,j} M_{ij}$, and its accuracy depends on the attention recall $r_i = \sum_j M_{ij} P_{ij}$.

\paragraph{Two-sided clustering.} SVG2~\cite{yang2025sparse} builds $M$ by running k-means on both sides, which partitions the key indices into $C_k$ clusters $K_c$ with means $\bar{k}_c$ and the query indices into $C_q$ clusters $Q_n$ with means $\bar{q}_n$, where query $i$ belongs to $Q_{n(i)}$. Replacing every member with its cluster mean gives the cluster-level logit $\bar{z}_{nc} = \bar{q}_n \bar{k}_c^\top / \sqrt{D}$, from which SVG2 estimates the attention mass that $Q_n$ places on $K_c$ as
\begin{equation}
\hat{P}_{nc} = \frac{|K_c|\, e^{\bar{z}_{nc}}}{\sum_{c'=1}^{C_k} |K_{c'}|\, e^{\bar{z}_{nc'}}}.
\label{eq:centroid-score}
\end{equation}
It then selects the top-$p$ set $\mathcal{S}_n$, the fewest key clusters whose scores sum to at least $p$, and lets every query in $Q_n$ attend to all keys of these clusters, so queries in one cluster share one mask row:
\begin{equation}
n(i) = n(i') \implies M_{i,:} = M_{i',:}.
\label{eq:shared}
\end{equation}
This fixes both \emph{which} key clusters and \emph{how many} keys each query attends to, as \Cref{fig:shared-vs-perquery}(a) shows. In exchange, the mask maps onto a block-sparse kernel: once the tokens are permuted by cluster, every selected pair of query and key clusters is a dense block~\cite{yang2025sparse}. Other clustering-based methods for video DiTs~\cite{luo2026attention,tan2026adacluster,zhou2026svgear} also share the selection.

\paragraph{Centroid compensation.} For each unselected key cluster $c$, SVG-EAR~\cite{zhou2026svgear} replaces the logits of the keys in $K_c$ with the per-query logit $z_{ic} = q_i \bar{k}_c^\top / \sqrt{D}$ and their values with the value mean $\bar{v}_c$, which gives
\begin{equation}
\hat{o}_i = \frac{\sum_{j} M_{ij}\, e^{s_{ij}} v_j + \sum_{c \notin \mathcal{S}_{n(i)}} |K_c|\, e^{z_{ic}}\, \bar{v}_c}{\sum_{j} M_{ij}\, e^{s_{ij}} + \sum_{c \notin \mathcal{S}_{n(i)}} |K_c|\, e^{z_{ic}}}.
\label{eq:comp}
\end{equation}
The compensation thus uses each query's own logits $z_{ic}$, whereas the selection uses the cluster-level logits $\bar{z}_{nc}$.

\subsection{Query Granularity and Accuracy}
\label{sec:accuracy}

\begin{figure*}[t]
\centering
\includegraphics[width=\textwidth]{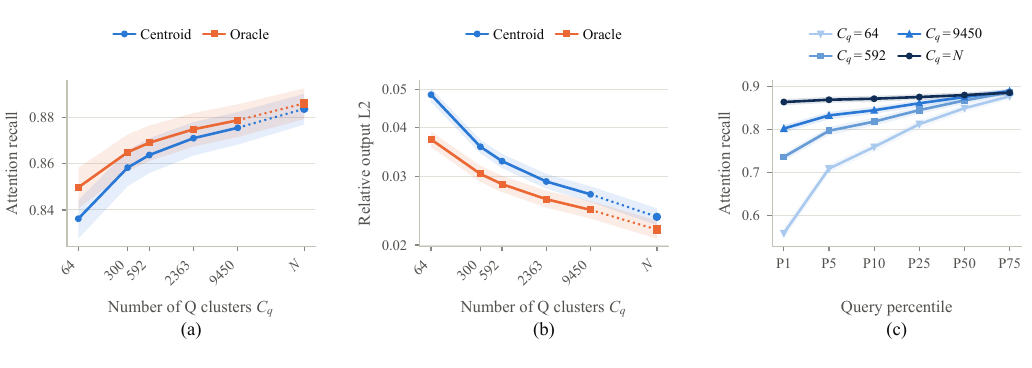}
\caption{\textbf{Accuracy versus query granularity.} All configurations share the key clusters ($C_k = 592$) and the target density ($\rho = 0.22$) and differ only in the number of query clusters $C_q$, where $C_q = N$ is per-query routing. (a) Mean recall under centroid and Oracle scoring. (b) Mean relative error of the compensated output. (c) Recall percentiles over the queries of a head under centroid scoring, averaged over heads.}
\label{fig:query-routing-granularity}
\end{figure*}

To measure how much accuracy the shared selection in \Cref{eq:shared} costs, we refine the query clusters at a fixed density. At $C_q = N$, every query forms its own cluster with $\bar{q}_{n(i)} = q_i$, so no selection is shared and we obtain per-query routing, as illustrated in \Cref{fig:shared-vs-perquery}(b).

We record $Q$, $K$, and $V$ along the dense sampling trajectories of Wan2.1-T2V-14B~\cite{wan2025wan} at 720p for eight of the augmented VBench prompts used by SVG2~\cite{huang2024vbench,yang2025sparse}. All configurations share the key clusters ($C_k = 592$) and run their own query k-means, with $p$ calibrated per head and configuration to a density of $\rho = 0.22$. To separate the error of the centroid estimate from the cost of sharing, we also select key clusters with Oracle scores, which replace $\hat{P}_{nc}$ with the true attention mass averaged over the queries of the cluster:
\begin{equation}
P^{\star}_{nc} = \frac{1}{|Q_n|} \sum_{i \in Q_n} \sum_{j \in K_c} P_{ij}.
\label{eq:oracle}
\end{equation}
We measure the recall $r_i$ and the relative output error $\|\hat{o}_i - o_i\|_2 / \|o_i\|_2$ of the compensated output in \Cref{eq:comp}.

\paragraph{Finer query clusters improve accuracy.} \Cref{fig:query-routing-granularity}(a, b) shows that refining the query clusters raises recall and lowers output error under both scoring rules, with per-query routing the most accurate. In a smaller cluster, fewer queries share one selection, which then fits each member better.

\paragraph{Oracle scores do not remove the cost of sharing.} The gap between Oracle and centroid scoring narrows as $C_q$ grows, since a smaller cluster is better represented by its mean. Yet at every shared granularity we test, even Oracle scoring is less accurate than per-query routing with centroid scores, so the remaining loss comes from the sharing itself rather than from the error of the centroid estimate.

\paragraph{The gain concentrates in the low-recall tail.} \Cref{fig:query-routing-granularity}(c) shows that refining the query clusters lifts the low recall percentiles the most. P1 rises from 0.560 at $C_q = 64$ to 0.865 under per-query routing, whereas above the median the curves nearly coincide, so per-query routing yields an almost flat profile. This follows from how selections are sized: per-query routing adds key clusters until each query's estimated mass reaches $p$, whereas a shared selection is sized by the cluster mean and under-covers the queries far from it, as the recall bars of \Cref{fig:shared-vs-perquery} show.

\subsection{Cost of Query Granularity}
\label{sec:cost}

\Cref{sec:accuracy} shows that finer query granularity improves accuracy. We now compare the routing cost of reaching it by two-sided clustering and by per-query routing. For one head in one attention call, we count the matrix multiplications relative to the $4 \rho N^2 D$ FLOPs that sparse attention spends on the logits and weighted values of the selected pairs, and leave out the key clustering that both share.

\paragraph{Refining query clusters.} Two-sided clustering adds the query k-means and the cluster-level scores of \Cref{eq:centroid-score}. With centroids warm-started from the previous denoising step~\cite{yang2025sparse}, the k-means runs only $I_q$ iterations per call, each comparing $N$ queries with $C_q$ centroids. Relative to sparse attention, this costs
\begin{equation}
R_{\text{two-sided}} = \frac{2 I_q N C_q D + 2 C_q C_k D}{4 \rho N^2 D} \approx \frac{I_q C_q}{2 \rho N},
\label{eq:cost-shared}
\end{equation}
where we drop the scoring term, which is only $C_k / (I_q N)$ of the k-means term.

\paragraph{Routing each query directly.} Per-query routing needs no query clustering and only scores each query against each key mean, which costs
\begin{equation}
R_{\text{per-query}} = \frac{2 N C_k D}{4 \rho N^2 D} = \frac{C_k}{2 \rho N}.
\label{eq:cost-per-query}
\end{equation}
For Wan2.1-T2V-14B at 720p ($N = 75{,}600$) with $C_k = 592$ and $\rho = 0.22$, this cost is 1.8\% of the sparse attention.

Dividing \Cref{eq:cost-shared} by \Cref{eq:cost-per-query} gives
\begin{equation}
\frac{R_{\text{two-sided}}}{R_{\text{per-query}}} \approx \frac{I_q C_q}{C_k},
\label{eq:cost-ratio}
\end{equation}
which depends neither on the video length nor on the density. Per-query routing is therefore cheaper whenever $I_q C_q > C_k$. With the $I_q = 2$ iterations that SVG2 uses and $C_k = 592$, this holds for $C_q > 296$, so every shared configuration in \Cref{fig:query-routing-granularity} except $C_q = 64$ costs more than per-query routing and is also less accurate.

\paragraph{Beyond matrix multiplications.} FLOPs alone do not determine the speed on GPUs, where tensor cores multiply a tile of typically 64 queries with a tile of keys, so a tile is efficient only when all of its queries attend to the same keys. Two-sided clustering meets this requirement by grouping the queries before they select keys, but refining these groups is costly: at fixed $C_k$, raising $C_q$ from 50 to 400 lifts the PSNR of SVG2~\cite{yang2025sparse} by 3.9~dB but cuts its speedup from $1.90\times$ to $1.25\times$. Under per-query routing the queries of a tile generally attend to different keys, so PQRS reverses this order: each query selects its own key clusters, and queries with the same selection then share a tile.

%% file: sec/4_method.tex
\section{Per-Query Routed Sparse Attention}
\label{sec:method}

\Cref{sec:routing} defines the output that PQRS computes for each query. \Cref{sec:execution} computes these outputs by traversing the key clusters, packing queries with identical selections into dense tasks, and merging the partial results of each query in a single pass, and \Cref{sec:stability} shows that this merging is numerically safe.

\subsection{Per-Query Routing and Compensation}
\label{sec:routing}

\begin{algorithm}[t]
  \caption{PQRS attention for one head.}
  \label{alg:pqrs}
  \begin{algorithmic}[1]
    \Require $Q, K, V \in \mathbb{R}^{N \times D}$; $C_k$; $p$; $f$
    \Ensure $\hat{o}_1, \dots, \hat{o}_N$
    \State $K_1, \dots, K_{C_k} \gets$ k-means on the keys \label{alg:cluster}
    \State Compute $|K_c|$, $\bar{k}_c$, and $\bar{v}_c$ for every cluster $c$ \label{alg:means}
    \State $Z \gets Q \bar{K}^\top / \sqrt{D}$ \Comment{$z_{ic}$ for all $i, c$} \label{alg:logits}
    \ForAll{queries $i$ in parallel}
      \State $\hat{P}_{ic} \gets |K_c|\, e^{z_{ic}} / \sum_{c'} |K_{c'}|\, e^{z_{ic'}}$ for every $c$ \label{alg:score}
      \State $\mathcal{S}_i \gets$ top clusters by $\hat{P}_{ic}$, added until \label{alg:select}
      \Statex \hspace{\algorithmicindent}\hspace{\algorithmicindent} $\sum_{c \in \mathcal{S}_i} \hat{P}_{ic} \ge p$ and $\sum_{c \in \mathcal{S}_i} |K_c| \ge fN$
      \State $(u_i, l_i) \gets \sum_{c \notin \mathcal{S}_i} |K_c|\, e^{z_{ic}} (\bar{v}_c, 1)$ \Comment{means} \label{alg:comp}
      \State $(u_i, l_i) \mathrel{+}= \sum_{c \in \mathcal{S}_i} \sum_{j \in K_c} e^{s_{ij}} (v_j, 1)$ \Comment{exact} \label{alg:exact}
      \State $\hat{o}_i \gets u_i / l_i$ \label{alg:output}
    \EndFor
  \end{algorithmic}
\end{algorithm}

\Cref{alg:pqrs} summarizes one sparse attention call of PQRS for one head in three steps.

\paragraph{Key clustering.} PQRS clusters only the keys. At every sparse attention call, it runs k-means on the keys of each head and stacks the key means as the rows of $\bar{K} \in \mathbb{R}^{C_k \times D}$.

\paragraph{Per-query routing.} Each query $i$ scores the key clusters with \Cref{eq:centroid-score}, using its own logit $z_{ic}$ in place of $\bar{z}_{nc}$, which gives $\hat{P}_{ic}$. It selects key clusters in descending order of $\hat{P}_{ic}$ until their scores sum to at least $p$ and, following the official implementation of SVG2~\cite{yang2025sparse}, until they cover at least $fN$ keys. Query $i$ attends to all keys of the selected set $\mathcal{S}_i$, which, unlike $\mathcal{S}_n$, depends on query $i$ alone.

\paragraph{Compensation.} PQRS keeps the centroid compensation of \Cref{eq:comp}, with $\mathcal{S}_i$ in place of $\mathcal{S}_{n(i)}$. Since routing has already computed $z_{ic}$ for every key cluster, the compensation adds only one weighted sum of the value means. In \Cref{alg:pqrs}, $u_i \in \mathbb{R}^D$ and $l_i$ accumulate the numerator and denominator of \Cref{eq:comp}, and $(v, 1)$ appends a one to a value vector so that both are accumulated together.

\subsection{Cluster-Major Execution}
\label{sec:execution}

\begin{figure*}[t]
  \centering
  \includegraphics[width=\textwidth]{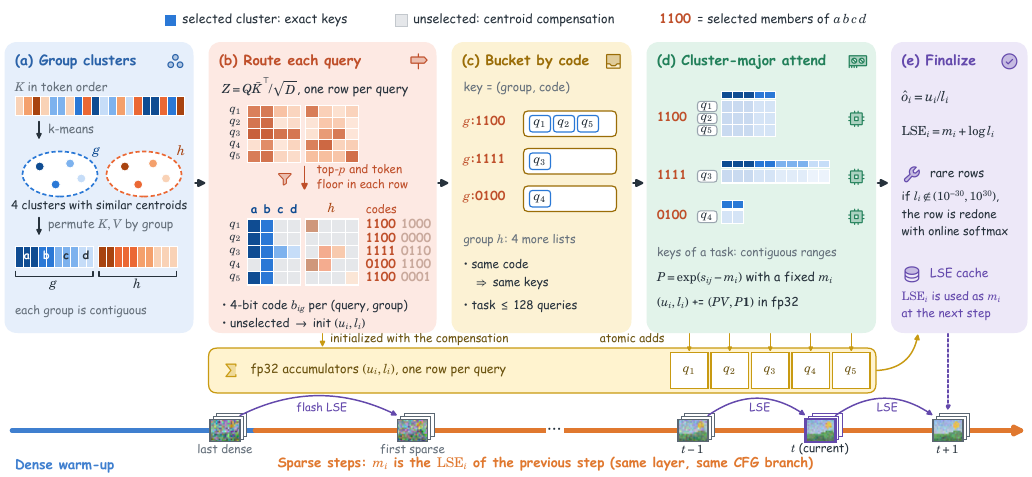}
  \caption{\textbf{Execution pipeline of PQRS} for one attention head. (a) Key clusters with similar centroids are grouped and stored contiguously. (b) Each query records its selection in each group as a 4-bit code and initializes its accumulators with the compensation. (c) The queries are bucketed by (group, code). (d) Each bucket runs as dense tasks that atomically add their partial results under a fixed offset $m_i$. (e) The rows are normalized, and rare rows outside the safe range are recomputed. Bottom: each row's LSE is its offset at the next step.}
  \label{fig:pipeline}
\end{figure*}

Since the selections differ across queries, PQRS computes Line~\ref{alg:exact} of \Cref{alg:pqrs} in cluster-major order, following KV-outer sparse attention kernels~\cite{lu2025moba,yan2026fsa,lai2026msa}: it traverses the key clusters and gathers, for each of them, the queries that selected it.

\paragraph{Grouping clusters.} A key cluster holds few keys, so a task over one cluster is short along the keys, yet each of its queries must still be gathered and have its partial output atomically merged. PQRS therefore executes groups of $G = 4$ clusters, so that a query pays this cost once per group it hits rather than once per selected cluster. PQRS groups them greedily by the cosine similarity of their key means. The grouping runs before routing, and $K$ and $V$ are permuted so that each group is contiguous in memory.


\paragraph{Routing each query.} Since $\mathcal{S}_i$ consists of the clusters whose scores reach a row-specific threshold, the router finds this threshold by bisection instead of sorting. It records the selection of query $i$ in group $g$ as a $G$-bit code $b_{ig}$ in member order and writes the compensation into the fp32 accumulators $(u_i, l_i)$ as their initial value.

\paragraph{Bucketing by code.} A query often selects only some members of a group it hits. Running the whole group for this query would count the unselected members twice, once exactly and once in the compensation, and masking them would waste tensor-core work. PQRS instead buckets the queries of each group by code, so all queries in a bucket select the same members and run as dense tasks of up to 128 queries over the contiguous key ranges of these members. In \Cref{fig:pipeline}, $q_1$, $q_2$, and $q_5$ select members $a$ and $b$ of group $g$ and share the code $1100$, so $g$ runs as three dense tasks without a union or a mask. A group can split its queries into up to $2^G - 1$ buckets, which leaves some tasks partially filled, and \Cref{sec:ablation} evaluates this trade-off over the group width.

\paragraph{Cluster-major attention.} The partial results of a query come from tasks that run on different SMs in no fixed order. Online softmax~\cite{dao2024flashattention} rescales the accumulated output whenever the running maximum grows and therefore combines them in sequence, and existing KV-outer kernels either recompute $QK^\top$ for the softmax statistics~\cite{yan2026fsa} or merge the stored partial results in another kernel~\cite{lai2026msa}. PQRS instead fixes a per-row offset $m_i$ before the kernel starts and subtracts it from every logit of row $i$, including $z_{ic}$, which scales the numerator and denominator of \Cref{eq:comp} by the same factor $e^{-m_i}$. The partial results then simply add: each task computes $P = \exp(s_{ij} - m_i)$ over its query--key pairs and adds $(PV, P\mathbf{1})$ to the accumulators of its queries.

\paragraph{Offsets from the previous step.} Let $\mathrm{LSE}^{(t)}_i$ be the natural logarithm of the denominator of \Cref{eq:comp} at step $t$, and let $\delta_i = \mathrm{LSE}^{(t)}_i - m_i$. The kernel accumulates $l_i = e^{\delta_i}$, and every term and partial sum of $l_i$ lies in $[0, l_i]$, so the numerical range of a row depends only on $\delta_i$. The ideal offset is therefore the current LSE, which is unknown before the kernel runs. Since the LSE is stable across denoising steps~\cite{xia2025adaspa}, PQRS sets $m_i = \mathrm{LSE}^{(t-1)}_i$ from the same layer and classifier-free guidance~\cite{ho2021classifierfree} branch, and the first sparse step uses the LSE that flash attention returns at the last dense step. A cached maximum logit would instead add to $\delta_i$ the gap between the LSE and the maximum up to $\ln N$. FlashDecoding++~\cite{hong2024flashdecoding} also fixes the offset in advance and recomputes on overflow, but uses one offset for all rows.

\paragraph{Finalization.} A final pass computes $\hat{o}_i = u_i / l_i$ and $\mathrm{LSE}^{(t)}_i = m_i + \ln l_i$ and caches the LSE for the next step. Rows with $l_i \notin (10^{-30}, 10^{30})$ are recorded in device memory and recomputed by single-row online softmax over the same selection and compensation, without host synchronization. Implementation details are given in the supplementary material.

\begin{figure}[t]
  \centering
  \includegraphics[width=\linewidth]{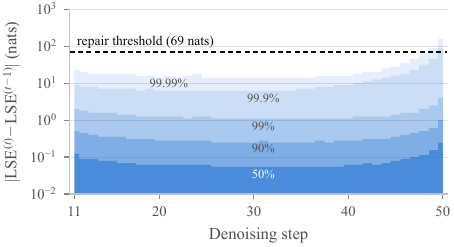}
  \caption{\textbf{The LSE changes little between consecutive steps.} Percentiles of $|\delta_i|$ over the sparse rows at each denoising step on Wan2.1-T2V-14B at 720p, pooled over two prompts. Rows above the dashed repair threshold, $l_i = 10^{\pm 30}$, are recomputed.}
  \label{fig:lse}
\end{figure}

\subsection{Numerical Safety of Cross-Step Offsets}
\label{sec:stability}

In exact arithmetic the offset cancels in $u_i / l_i$ and only sets the range of the fp32 accumulators. Within the range accepted by the final pass, $|\delta_i| < \ln 10^{30} \approx 69.08$, the accumulators stay far from overflow, and the mass lost to underflow stays below the bf16 rounding error of the output. Other rows are recomputed, so the output is accurate for any offset, and a good offset only means fewer repaired rows.

\paragraph{Why the previous LSE suffices.} The tolerance needs only a coarse prediction. For a fixed selection, the LSE of a row changes, to first order, by the attention-weighted average of the changes in its logits, so it drifts far only when these logits move together, and refreshing the cache at every step keeps the error from accumulating.

\paragraph{Stability in practice.} On Wan2.1-T2V-14B, \Cref{fig:lse} shows that fewer than $0.01\%$ of the rows cross the repair threshold, almost all of them in the last two steps. It indicates that the LSE is stable enough to reuse across steps.

%% file: sec/5_exp.tex
\begin{table*}[t]
  \centering
  \small
  \setlength{\tabcolsep}{7pt}
  \setlength{\aboverulesep}{0pt}
  \setlength{\belowrulesep}{0pt}
  \renewcommand{\arraystretch}{1}
  \caption{\textbf{Quality and efficiency on four models.}
  Among sparse methods, the best result is in \textbf{bold} and the second best is \underline{underlined}.}
  \label{tab:main}
  \begin{tabular}{l|l|cccc|ccc}
    \toprule
    \multirow{2}{*}{Model} & \multirow{2}{*}{Method} & \multicolumn{4}{c|}{Quality} & \multicolumn{3}{c}{Efficiency} \\
    \cmidrule(lr){3-6} \cmidrule(l){7-9}
    & & PSNR$\uparrow$ & SSIM$\uparrow$ & LPIPS$\downarrow$ & VBench$\uparrow$ & Density & Latency$\downarrow$ & Speedup$\uparrow$ \\
    \midrule
    \multirow{7}{*}{\shortstack[l]{Wan2.1-T2V-14B\\\textit{(720P, 81 Frames)}}}
      & Dense   & --    & --    & --    & 83.64\% & 100\%  & 1682s & 1.00$\times$ \\
    & SVG     & 22.17 & 0.769 & 0.210 & \underline{83.77\%} & 40.7\% & 1053s & 1.60$\times$ \\
    & SVG2    & 25.73 & 0.858 & 0.138 & 83.54\% & 31.4\% & 1091s & 1.54$\times$ \\
    & SVOO    & 26.02 & 0.866 & 0.130 & 83.58\% & 35.8\% & 1137s & 1.48$\times$ \\
    & PISA    & 24.05 & 0.824 & 0.164 & \textbf{83.78\%} & 35.5\% & 1107s & 1.52$\times$ \\
    & SVG-EAR & \underline{27.19} & \underline{0.884} & \underline{0.116} & 83.58\% & 21.7\% & \underline{1032s} & \underline{1.63}$\times$ \\
    & \cellcolor{pqrsblue}\textbf{PQRS} & \cellcolor{pqrsblue}\textbf{28.86} & \cellcolor{pqrsblue}\textbf{0.909} & \cellcolor{pqrsblue}\textbf{0.094} & \cellcolor{pqrsblue}83.70\% & \cellcolor{pqrsblue}22.0\% & \cellcolor{pqrsblue}\textbf{960s} & \cellcolor{pqrsblue}\textbf{1.75}$\times$ \\
    \midrule
    \multirow{7}{*}{\shortstack[l]{Wan2.1-I2V-14B\\\textit{(720P, 81 Frames)}}}
      & Dense   & --    & --    & --    & 83.74\% & 100\%  & 1362s & 1.00$\times$ \\
    & SVG     & 23.98 & 0.790 & 0.183 & 83.50\% & 40.7\% & 858s  & 1.59$\times$ \\
    & SVG2    & 26.59 & 0.857 & 0.140 & 83.47\% & 30.4\% & 878s  & 1.55$\times$ \\
    & SVOO    & 27.27 & 0.873 & 0.128 & 83.57\% & 33.2\% & 919s  & 1.48$\times$ \\
    & PISA    & 26.85 & 0.869 & 0.130 & \underline{83.69\%} & 35.5\% & 866s  & 1.57$\times$ \\
    & SVG-EAR & \underline{28.06} & \underline{0.889} & \underline{0.117} & 83.59\% & 21.2\% & \underline{838s} & \underline{1.62}$\times$ \\
    & \cellcolor{pqrsblue}\textbf{PQRS} & \cellcolor{pqrsblue}\textbf{29.97} & \cellcolor{pqrsblue}\textbf{0.917} & \cellcolor{pqrsblue}\textbf{0.097} & \cellcolor{pqrsblue}\textbf{83.74\%} & \cellcolor{pqrsblue}21.6\% & \cellcolor{pqrsblue}\textbf{781s} & \cellcolor{pqrsblue}\textbf{1.74}$\times$ \\
    \midrule
    \multirow{7}{*}{\shortstack[l]{Wan2.2-T2V-A14B\\\textit{(720P, 81 Frames)}}}
      & Dense   & --    & --    & --    & 84.26\% & 100\%  & 1424s & 1.00$\times$ \\
    & SVG     & 17.66 & 0.594 & 0.328 & 84.12\% & 40.7\% & 926s  & 1.54$\times$ \\
    & SVG2    & 21.15 & 0.744 & 0.222 & 84.24\% & 33.6\% & 986s  & 1.44$\times$ \\
    & SVOO    & 20.87 & 0.735 & 0.227 & 84.21\% & 39.3\% & 1037s & 1.37$\times$ \\
    & PISA    & 19.13 & 0.665 & 0.276 & \textbf{84.32\%} & 35.5\% & 931s  & 1.53$\times$ \\
    & SVG-EAR & \underline{22.25} & \underline{0.779} & \underline{0.194} & 84.19\% & 22.6\% & \underline{923s} & \underline{1.54}$\times$ \\
    & \cellcolor{pqrsblue}\textbf{PQRS} & \cellcolor{pqrsblue}\textbf{23.47} & \cellcolor{pqrsblue}\textbf{0.814} & \cellcolor{pqrsblue}\textbf{0.170} & \cellcolor{pqrsblue}\underline{84.26\%} & \cellcolor{pqrsblue}22.5\% & \cellcolor{pqrsblue}\textbf{856s} & \cellcolor{pqrsblue}\textbf{1.66}$\times$ \\
    \midrule
    \multirow{7}{*}{\shortstack[l]{HunyuanVideo-T2V-13B\\\textit{(720P, 129 Frames)}}}
      & Dense   & --    & --    & --    & 84.45\% & 100\%  & 1701s & 1.00$\times$ \\
    & SVG     & 27.34 & 0.892 & 0.125 & 84.37\% & 36.6\% & 961s  & 1.77$\times$ \\
    & SVG2    & 31.09 & 0.933 & 0.084 & 84.30\% & 26.8\% & 961s  & 1.77$\times$ \\
    & SVOO    & 29.95 & 0.922 & 0.093 & 84.25\% & 25.0\% & \underline{881s} & \underline{1.93}$\times$ \\
    & PISA    & 28.27 & 0.909 & 0.109 & \underline{84.46\%} & 35.5\% & 1085s & 1.57$\times$ \\
    & SVG-EAR & \underline{32.95} & \underline{0.948} & \underline{0.068} & 84.38\% & 19.8\% & 929s  & 1.83$\times$ \\
    & \cellcolor{pqrsblue}\textbf{PQRS} & \cellcolor{pqrsblue}\textbf{35.44} & \cellcolor{pqrsblue}\textbf{0.961} & \cellcolor{pqrsblue}\textbf{0.053} & \cellcolor{pqrsblue}\textbf{84.48\%} & \cellcolor{pqrsblue}19.6\% & \cellcolor{pqrsblue}\textbf{809s} & \cellcolor{pqrsblue}\textbf{2.10}$\times$ \\
    \bottomrule
  \end{tabular}
\end{table*}

\section{Experiments}
\label{sec:experiments}

\subsection{Setup}
\label{sec:setup}

\paragraph{Models.}
We evaluate PQRS on four widely used video generation models: Wan2.1-T2V-14B, Wan2.2-T2V-A14B~\cite{wan2025wan} and HunyuanVideo-T2V-13B~\cite{kong2025hunyuanvideo} for text-to-video (T2V) generation, and Wan2.1-I2V-14B~\cite{wan2025wan} for image-to-video (I2V) generation.

\paragraph{Metrics.}
For quality, we measure the similarity between the videos generated with sparse and dense attention by PSNR, SSIM~\cite{wang2004ssim} and LPIPS~\cite{zhang2018unreasonable}, and report the VBench score~\cite{huang2024vbench} averaged over subject consistency, background consistency, motion smoothness, aesthetic quality and imaging quality. For efficiency, we report the attention density, the fraction of query--key pairs computed exactly in the sparse attention calls, together with the end-to-end latency and the speedup over dense attention.

\paragraph{Datasets.}
Following prior work~\cite{yang2025sparse,zhou2026svgear,luo2026attention}, we randomly sample 50 of the augmented VBench prompts~\cite{huang2024vbench} for T2V and 50 image--prompt pairs from VBench++~\cite{huang2025vbenchpp} for I2V.

\paragraph{Baselines.}
We compare PQRS with state-of-the-art training-free sparse attention methods for video generation: SVG~\cite{xi2025sparse}, SVG2~\cite{yang2025sparse}, SVG-EAR~\cite{zhou2026svgear}, SVOO~\cite{luo2026attention} and PISA~\cite{li2026pisa}. All baselines use their official configurations, except for the shared warm-up below and the density hyperparameter. \Cref{tab:main} lists the resulting densities.

\paragraph{Implementation.}
PQRS uses $C_k = 592$ key clusters, a routing threshold of $p = 0.848$ and a routing floor of $f = 0.10$ on all four models. Following~\cite{zhou2026svgear,li2026pisa}, every method computes dense attention in the first 20\% of denoising steps and in the first block of each transformer stack. All methods, including dense attention with PyTorch SDPA, share the same non-attention operators, so latency differences come from attention alone. Videos are generated at 720p with the official sampling configuration of each model on a single NVIDIA H100 GPU, and further details are given in the supplementary material.


\subsection{Main Results}
\label{sec:main-results}

\Cref{tab:main} compares PQRS with five sparse attention baselines on four video diffusion models, covering both text-to-video and image-to-video generation. PQRS consistently achieves the best PSNR, SSIM, and LPIPS on all four models, surpassing SVG-EAR, which runs at a comparable density, as well as SVG, SVG2, SVOO, and PISA, which compute more attention. Since each query selects key clusters according to its own attention distribution, PQRS retains the attention mass that a shared selection misses, and the generated videos stay closest to the dense outputs. On the reference-free VBench metric, PQRS remains on par with dense attention across all four models, and the qualitative comparisons in \Cref{fig:teaser} further confirm that it preserves the visual quality of the dense outputs.

PQRS also delivers the highest inference efficiency on every model, reaching 1.66--2.10$\times$ end-to-end speedup. Although per-query routing produces an irregular selection pattern, the cluster-major kernel in \Cref{sec:execution} executes it efficiently, so the finer selection does not come at the cost of speed. As a result, PQRS improves fidelity and inference speed at the same time on all evaluated models, achieving a better quality--efficiency trade-off than existing sparse attention methods.

\subsection{Ablation Study}
\label{sec:ablation}

\begin{figure}[t]
  \centering
  \includegraphics[width=\linewidth]{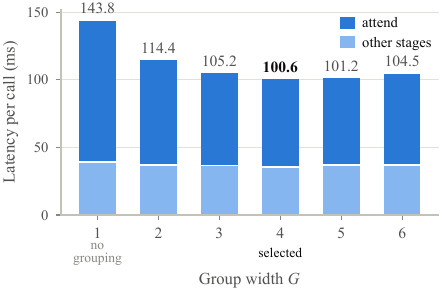}
  \caption{\textbf{Latency versus group width $G$.} Each bar is one sparse attention call over all $40$ heads of Wan2.1-T2V-14B, averaged over $3$ prompts, $3$ denoising steps and $39$ layers. \emph{Other stages} covers cluster, route, build and finalize.}
  \label{fig:group-width}
\end{figure}

\paragraph{Group width.}
We sweep the group width $G$ from $1$ to $6$, keep every other setting at its main-table value, and time all widths in one process on the same Q, K and V tensors in alternating order. Neither the routing rule nor the estimator depends on $G$, so all widths select the same clusters and their outputs differ only in floating-point summation order.

As shown in \Cref{fig:group-width}, $G=4$ is the fastest width, and removing grouping ($G=1$) makes a call $1.4\times$ slower, almost entirely in attend, as the stages outside it vary by less than $4$~ms across the sweep. A group is built from similar clusters, which queries tend to select together: at $G=4$, a query that touches a group selects $2.3$ of its four clusters, and one task serves them with a single Q gather and a single merge, rather than one of each per cluster. A wider group saves more of both, but only queries that select the same subset of the group can share a task, and $G$ clusters have $2^G-1$ non-empty subsets. At $G=6$ an average task fills only $77\%$ of its $128$ query rows and computes the rest anyway, so beyond $G=4$ the emptier tasks outweigh the saving.

\paragraph{Selection versus compensation.}
SVG-EAR~\cite{zhou2026svgear} applies the same centroid compensation as PQRS, but to a selection shared within each query cluster. To separate the two components, we cross shared or per-query selection with compensation off or on and keep every other setting at its main-table value. The shared cells cluster the queries into $C_q=C_k=592$ clusters, select key clusters for each query cluster with the SVG2 rule~\cite{yang2025sparse}, and recalibrate $p$ to the density of per-query selection.

As shown in \Cref{tab:compensation}, each component improves every metric on its own, and per-query selection beats shared selection by $0.9$\,dB of PSNR even without compensation. The two gains also reinforce each other: compensation adds $1.7$\,dB on top of shared selection but $3.1$\,dB on top of per-query selection, and SSIM and LPIPS follow the same pattern. A likely reason is that the centroid estimate works best for clusters that carry little of a query's attention. Per-query selection leaves the estimate only each query's light tail, whereas a shared selection under-covers the queries far from the cluster mean, as shown in \Cref{sec:accuracy}, and leaves some of their heavily attended clusters to the estimate.

\begin{table}[ht]
\centering
\small
\setlength{\tabcolsep}{5pt}
\caption{\textbf{Selection granularity $\times$ centroid compensation.} Results on Wan2.1-T2V-14B, where shared selection uses $C_q=592$ query clusters and the last row is PQRS.}
\label{tab:compensation}
\begin{tabular}{@{}lccccc@{}}
\toprule
Selection & Comp. & Density & PSNR$\uparrow$ & SSIM$\uparrow$ & LPIPS$\downarrow$ \\
\midrule
Shared    & \xmark & 22.5\% & 24.91 & 0.842 & 0.151 \\
Shared    & \cmark & 21.9\% & 26.61 & 0.874 & 0.123 \\
Per-query & \xmark & 22.4\% & 25.81 & 0.861 & 0.134 \\
Per-query & \cmark & 22.0\% & \textbf{28.86} & \textbf{0.909} & \textbf{0.094} \\
\bottomrule
\end{tabular}
\end{table}

%% file: sec/6_conclusion.tex
\section{Conclusion}
\label{sec:conclusion}

We propose PQRS, a training-free sparse attention method for video DiTs. Since sharing one selection across each query cluster limits accuracy, PQRS clusters only the keys, lets every query select its own key clusters, and approximates the unselected ones by their centroids. A cluster-major kernel packs queries with identical selections into dense tasks and merges their partial results in a single pass. On four video generation models, PQRS is faster and more faithful to dense attention than all five baselines, reaching 1.66--2.10$\times$ end-to-end speedup.

%% file: sec/X_suppl.tex
\clearpage

\appendix
\renewcommand{\topfraction}{0.85}
\renewcommand{\textfraction}{0.1}

\section{Implementation Details of PQRS}
\label{sec:supp-impl}

\paragraph{Kernels.} The attend and compensation kernels are written with CuTe and CUTLASS for H100, selection and task construction in CUDA, and the remaining stages in Triton. All inputs are BF16.

\paragraph{Clustering and grouping.} Each sparse call clusters the keys of each head, after QK normalization and RoPE, into $C_k = 592$ clusters with the k-means kernel of SVG2~\cite{yang2025sparse}. The first sparse call runs 50 iterations. Each later call runs 2 iterations from the centroids of the previous call of the same layer, in either guidance branch. Grouping is greedy: the largest unassigned cluster opens a group, which then repeatedly takes the unassigned cluster with the largest summed cosine similarity to its centroids until it holds $G = 4$ clusters. Keys and values are permuted so that each group is contiguous in memory, and a task reads the selected members of a group as at most two contiguous segments.

\paragraph{Routing and execution.} One matrix multiplication scores all queries against all key centroids. Selection assigns one warp to each query, finds the threshold of \Cref{sec:routing} by bisection over the scores held in registers, and writes the 4-bit code of every group with a warp ballot. Task construction then buckets the queries of each head and band of query rows by group and code. A separate kernel first writes the compensation into the FP32 accumulators. The persistent attend kernel then loads keys and values with TMA, computes with wgmma and adds its partial results to the accumulators with vectorized atomic reductions.

\paragraph{Offsets and row repair.} At the last dense step, PQRS calls the FlashAttention kernel behind SDPA directly. It returns the same output together with the LSE, which gives the offsets of the first sparse step. Rows outside the safe range are appended to a list in device memory. The repair kernel reads the length of this list on the device and returns at once when the list is empty, so the host never synchronizes. As \Cref{tab:supp-repair} shows, repair is rare. On the two prompts of \Cref{fig:lse}, almost all repaired rows underflow, and they concentrate in the last two layers and the last two denoising steps.

\begin{table}[ht]
  \centering
  \caption{\textbf{Row repair on the 50 evaluation videos.} Rows is the fraction of (head, row) pairs in sparse calls that are repaired, and Calls is the fraction of sparse calls with at least one repaired row.}
  \label{tab:supp-repair}
  \small
  \setlength{\tabcolsep}{5pt}
  \begin{tabular}{lrr}
    \toprule
    Model & Rows & Calls \\
    \midrule
    Wan2.1-T2V-14B & $3.2 \times 10^{-5}$ & 8.7\% \\
    Wan2.1-I2V-14B & $3.9 \times 10^{-5}$ & 9.1\% \\
    Wan2.2-T2V-A14B & $3.3 \times 10^{-5}$ & 9.3\% \\
    HunyuanVideo-T2V-13B & 11 rows in total & 3 calls in total \\
    \bottomrule
  \end{tabular}
\end{table}

\paragraph{Wan2.2.} The two experts have separate layers, and only the expert active at the last dense step receives the dense LSE. In the other expert, the first sparse call of each layer therefore computes its offsets exactly with a pre-pass over the selected keys and runs k-means for 50 iterations. All later calls use the cached LSE.

\paragraph{HunyuanVideo.} HunyuanVideo attends jointly over the video tokens and 256 text tokens, of which the trailing ones are padding. PQRS sparsifies only the attention of video queries to video keys, and the density counts only this part. Each video query also attends to all valid text keys with one FlashAttention call, and the two parts are merged exactly through their LSEs. Text queries attend densely to all valid keys and padding rows only to padding, as in the official SVG2 code. The offsets of the first sparse step are the dense LSEs minus the contribution of the valid text keys.

\section{Experimental Setup}
\label{sec:supp-setup}

\begin{table}[t]
  \centering
  \caption{\textbf{VBench scores per dimension (\%).} Subj., Back., Motion, Aesth.\ and Imag.\ are subject consistency, background consistency, motion smoothness, aesthetic quality and imaging quality, and Avg.\ is their mean as reported in \Cref{tab:main}.}
  \label{tab:supp-vbench}
  \footnotesize
  \setlength{\tabcolsep}{5.5pt}
  \begin{tabular}{lcccccc}
    \toprule
    Method & Subj. & Back. & Motion & Aesth. & Imag. & Avg. \\
\midrule
    \multicolumn{7}{l}{\textit{Wan2.1-T2V-14B}} \\
    Dense & 93.86 & 95.73 & 98.62 & 62.22 & 67.76 & 83.64 \\
    SVG & 93.80 & 95.77 & 98.66 & 62.63 & 68.01 & 83.77 \\
    SVG2 & 93.71 & 95.69 & 98.60 & 62.11 & 67.56 & 83.54 \\
    SVOO & 93.86 & 95.55 & 98.62 & 62.12 & 67.73 & 83.58 \\
    PISA & 93.88 & 95.91 & 98.64 & 62.47 & 67.99 & 83.78 \\
    SVG-EAR & 93.72 & 95.51 & 98.63 & 62.18 & 67.88 & 83.58 \\
    \rowcolor{pqrsblue} \textbf{PQRS} & 93.81 & 95.92 & 98.63 & 62.43 & 67.70 & 83.70 \\
    \midrule
    \multicolumn{7}{l}{\textit{Wan2.1-I2V-14B}} \\
    Dense & 94.57 & 95.46 & 98.59 & 60.60 & 69.47 & 83.74 \\
    SVG & 94.08 & 94.86 & 98.56 & 60.24 & 69.73 & 83.50 \\
    SVG2 & 94.44 & 95.29 & 98.51 & 59.86 & 69.23 & 83.47 \\
    SVOO & 94.50 & 95.33 & 98.54 & 60.03 & 69.46 & 83.57 \\
    PISA & 94.55 & 95.40 & 98.57 & 60.43 & 69.52 & 83.69 \\
    SVG-EAR & 94.50 & 95.33 & 98.57 & 60.15 & 69.41 & 83.59 \\
    \rowcolor{pqrsblue} \textbf{PQRS} & 94.56 & 95.49 & 98.57 & 60.63 & 69.46 & 83.74 \\
    \midrule
    \multicolumn{7}{l}{\textit{Wan2.2-T2V-A14B}} \\
    Dense & 93.12 & 95.20 & 98.16 & 64.36 & 70.45 & 84.26 \\
    SVG & 93.12 & 95.20 & 98.18 & 63.69 & 70.41 & 84.12 \\
    SVG2 & 93.22 & 95.05 & 98.16 & 64.23 & 70.55 & 84.24 \\
    SVOO & 93.30 & 95.21 & 98.18 & 64.01 & 70.37 & 84.21 \\
    PISA & 93.25 & 95.38 & 98.21 & 64.33 & 70.41 & 84.32 \\
    SVG-EAR & 93.05 & 95.15 & 98.19 & 64.41 & 70.16 & 84.19 \\
    \rowcolor{pqrsblue} \textbf{PQRS} & 93.23 & 95.25 & 98.17 & 64.19 & 70.48 & 84.26 \\
    \midrule
    \multicolumn{7}{l}{\textit{HunyuanVideo-T2V-13B}} \\
    Dense & 94.11 & 96.00 & 99.38 & 62.26 & 70.49 & 84.45 \\
    SVG & 94.15 & 96.05 & 99.37 & 62.01 & 70.25 & 84.37 \\
    SVG2 & 93.97 & 95.93 & 99.35 & 62.23 & 70.03 & 84.30 \\
    SVOO & 94.02 & 95.92 & 99.35 & 62.10 & 69.88 & 84.25 \\
    PISA & 94.15 & 96.00 & 99.37 & 62.35 & 70.40 & 84.46 \\
    SVG-EAR & 94.04 & 96.09 & 99.37 & 62.12 & 70.30 & 84.38 \\
    \rowcolor{pqrsblue} \textbf{PQRS} & 94.12 & 96.06 & 99.38 & 62.32 & 70.49 & 84.48 \\
    \bottomrule
  \end{tabular}
\end{table}

\begin{table*}[ht]
  \centering
  \caption{\textbf{Sampling configurations.} Dense steps and dense blocks use dense attention for every method. $N$ counts video tokens. Euler denotes the flow-matching Euler sampler.}
  \label{tab:supp-models}
  \small
  \setlength{\tabcolsep}{4pt}
  \begin{tabular}{lccccccc}
    \toprule
    Model & Frames (fps) & $N$ & Steps & Sampler & Guidance & Dense steps & Dense blocks \\
    \midrule
    Wan2.1-T2V-14B & 81 (16) & 75,600 & 50 & UniPC, shift 5.0 & CFG 5.0 & 1--10 & first of 40 \\
    Wan2.1-I2V-14B & 81 (16) & 75,600 & 40 & UniPC, shift 5.0 & CFG 5.0 & 1--8 & first of 40 \\
    Wan2.2-T2V-A14B & 81 (16) & 75,600 & 40 & UniPC, shift 12.0 & CFG 4.0 / 3.0 & 1--8 & first per expert \\
    HunyuanVideo-T2V-13B & 129 (24) & 118,800 & 50 & Euler, shift 7.0 & embedded 6.0 & 1--10 & first of each stream type \\
    \bottomrule
  \end{tabular}
\end{table*}

\begin{table*}[ht]
  \centering
  \caption{\textbf{Baseline hyperparameters.} Each baseline uses the same density hyperparameter on all four models, except where noted, and its official values for all other hyperparameters.}
  \label{tab:supp-knobs}
  \small
  \setlength{\tabcolsep}{5pt}
  \begin{tabular}{lll>{\raggedright\arraybackslash}p{0.41\linewidth}}
    \toprule
    Method & Density hyperparameter & Official value & Other hyperparameters (official) \\
    \midrule
    SVG & sparsity 0.37 & 0.30 (HunyuanVideo 0.25) & 64 sampled query rows, FlexAttention \\
    SVG2 & top-$p$ 0.90 & 0.90 & $C_q / C_k = 300 / 1000$ (HunyuanVideo $400 / 1000$), k-means 50 / 2 iterations, floor 0.10 \\
    SVG-EAR & top-$p$ 0.80 & 0.85 & as SVG2, floor 0.05, $\gamma = 1$ \\
    SVOO & top-$p$ 0.90 (HunyuanVideo 0.88) & 0.90 (HunyuanVideo 0.88) & $C_q / C_k = 256 / 1024$, k-means 2 / 2 iterations, per-model schedule and profile \\
    PISA & density 0.355 & 0.125 & block size 64, no bias term, sink: none (Wan), last block (HunyuanVideo) \\
    \rowcolor{pqrsblue} \textbf{PQRS} & $p$ 0.848 & -- & $C_k = 592$, floor $f = 0.10$, k-means 50 / 2 iterations, $G = 4$ \\
    \bottomrule
  \end{tabular}
\end{table*}

\paragraph{Sampling.} The sampling configurations in \Cref{tab:supp-models} follow the official SVG inference scripts for Wan2.1 and HunyuanVideo and the official configuration of Wan2.2, whose two guidance scales belong to its two experts. HunyuanVideo uses embedded guidance and thus one transformer pass per step. On Wan2.2, every method offloads the two experts to the CPU and keeps the text encoder and the VAE on the GPU.

\paragraph{Prompts and metrics.} The augmented VBench prompts are the versions that the VBench team rewrote with prompt optimization. The I2V images are cropped to 16:9. PSNR, SSIM and LPIPS compare each frame with the same frame of the dense video generated from the same prompt and seed, and are averaged over frames and then over videos. SSIM uses an $11 \times 11$ uniform window, and LPIPS uses the VGG backbone. VBench 0.1.5 runs in its custom-input mode with the default preprocessing.

\section{Baseline Configurations}
\label{sec:supp-baselines}

\paragraph{Implementations.} Every baseline uses its official attention kernels. SVG~\cite{xi2025sparse}, SVG2~\cite{yang2025sparse}, SVG-EAR~\cite{zhou2026svgear} and SVOO~\cite{luo2026attention} run in an open-source framework that calls these kernels, and we checked its reimplementation of the other steps against the official code. PISA~\cite{li2026pisa} is called directly, as described below.

\paragraph{Hyperparameters.} In \Cref{tab:supp-knobs}, three density hyperparameters differ from the official values. SVG-EAR uses a lower top-$p$ to match the density of PQRS, and SVG and PISA are set to an end-to-end speedup of about $1.5\times$ on Wan2.1-T2V-14B, comparable to the other baselines.

\paragraph{Shared protocol.} Wherever attention is dense, in the dense baseline and in the dense steps and blocks of every method, it runs PyTorch SDPA with the FlashAttention-2 backend, as in the official SVG code. For the other operators, all methods including the dense baseline use the fused Triton kernels of SVG and SVOO, in which we fixed a LayerNorm bug that included zero-padded columns in the variance. The speedups reported by the SVG family use the original Diffusers operators for the dense pipeline and these fused kernels for the sparse one, so they also count the gain of kernel fusion. Our speedups use the fused kernels on both sides.

\paragraph{Adaptations.} Three baseline configurations needed adaptation.
\begin{itemize}
\item \emph{SVG-EAR on HunyuanVideo.} The official code passes the zero-filled centroids of the padding text tokens to the compensation, which adds a term absent from dense attention to the denominator of every video row. We exclude the prompt and padding clusters from the compensation and compute the prompt columns exactly.
\item \emph{PISA.} PISA provides no inference script for Wan or HunyuanVideo, so we call its kernel directly. Because the kernel takes no mask, on HunyuanVideo we remove the padding tokens and, as its documentation recommends for MMDiT models, use the last block as the attention sink. When the valid text spans more than one block, only the last block is guaranteed to be exact, which slightly disadvantages PISA.
\item \emph{SVOO on Wan2.2.} The official profile of SVOO for Wan2.2 was collected with a different shift and guidance scale. We reuse it because the profile only sets a per-head lower bound on the budget, which is clipped to $[0.20, 0.22]$ and lies well below the density of SVOO in \Cref{tab:main}.
\end{itemize}

\begin{table}[t]
  \centering
  \caption{\textbf{Results of \Cref{fig:supp-e10}.} The setting is $p$ for PQRS and top-$p$ for the baselines, and the middle row of each method is its setting in \Cref{tab:main}. Speedups are over dense attention (1682~s).}
  \label{tab:supp-e10}
  \footnotesize
  \setlength{\tabcolsep}{2.4pt}
  \begin{tabular}{lccccccc}
    \toprule
    Method & Setting & Density & Latency & Speedup & PSNR & SSIM & LPIPS \\
    \midrule
    SVG2 & 0.80 & 22.3\% & 994s & 1.69$\times$ & 24.66 & 0.836 & 0.157 \\
    SVG2 & 0.90 & 31.4\% & 1091s & 1.54$\times$ & 25.73 & 0.858 & 0.138 \\
    SVG2 & 0.95 & 50.8\% & 1268s & 1.33$\times$ & 26.67 & 0.875 & 0.123 \\
    \midrule
    SVG-EAR & 0.75 & 19.3\% & 1017s & 1.65$\times$ & 27.00 & 0.880 & 0.120 \\
    SVG-EAR & 0.80 & 21.7\% & 1032s & 1.63$\times$ & 27.19 & 0.884 & 0.116 \\
    SVG-EAR & 0.85 & 25.2\% & 1081s & 1.56$\times$ & 27.31 & 0.886 & 0.114 \\
    \midrule
    \rowcolor{pqrsblue} \textbf{PQRS} & 0.75 & 17.6\% & 933s & 1.80$\times$ & 28.39 & 0.903 & 0.100 \\
    \rowcolor{pqrsblue} \textbf{PQRS} & 0.848 & 22.0\% & 960s & 1.75$\times$ & 28.86 & 0.909 & 0.094 \\
    \rowcolor{pqrsblue} \textbf{PQRS} & 0.90 & 26.0\% & 989s & 1.70$\times$ & 29.43 & 0.916 & 0.088 \\
    \bottomrule
  \end{tabular}
\end{table}

\section{Additional Results}
\label{sec:supp-results}

\subsection{VBench Dimensions}

\Cref{tab:supp-vbench} breaks the VBench average of \Cref{tab:main} into its five dimensions. PQRS stays close to dense attention in all five dimensions on all four models.

\subsection{Quality at Different Budgets}
\label{sec:supp-e10}

\begin{figure*}[ht]
  \centering
  \includegraphics[width=\linewidth]{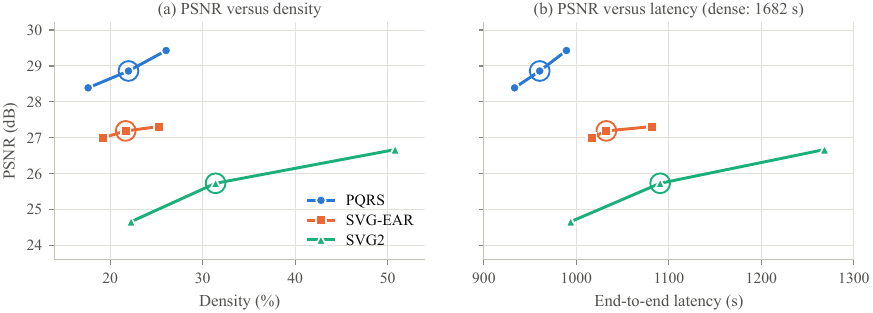}
  \caption{\textbf{Fidelity versus budget on Wan2.1-T2V-14B.} Each method runs at three settings, and circles mark the settings of \Cref{tab:main}. (a) PSNR versus density. (b) PSNR versus end-to-end latency.}
  \label{fig:supp-e10}
\end{figure*}

\Cref{fig:supp-e10} and \Cref{tab:supp-e10} add two settings for each of PQRS, SVG-EAR and SVG2 on Wan2.1-T2V-14B, one of them the official top-$p$ of SVG-EAR. In both panels of \Cref{fig:supp-e10}, PQRS at $p = 0.75$ lies above and to the left of every baseline point, and the paired bootstrap 95\% interval of its PSNR gain over each of them excludes zero.

\subsection{Per-Call Latency Breakdown}
\label{sec:supp-e11}

\begin{table*}[ht]
  \centering
  \caption{\textbf{Latency of one sparse attention call on Wan2.1-T2V-14B.} Times are in milliseconds per call over all 40 heads, averaged over 351 calls. Speedups are means of per-call ratios with bootstrap 95\% intervals. \emph{Outside attend} is the wall time minus the attend kernel, with its share of the wall time in parentheses. The reclustering row times an SVOO call that recomputes its clusters, and the amortized row spreads one such call over 40 calls.}
  \label{tab:supp-e11}
  \small
  \setlength{\tabcolsep}{4pt}
  \begin{tabular}{lccccc>{\raggedright\arraybackslash}p{0.29\linewidth}}
    \toprule
    Method & Wall time & Speedup & Density & Attend & Outside attend & Largest stages outside attend \\
    \midrule
    Dense (SDPA) & 329.1 & 1.00$\times$ & 100\% & 329.1 & -- & -- \\
    SVG & 123.3 & 2.67$\times$ [2.66, 2.68] & 40.7\% & 112.4 & 10.9 (8.8\%) & head profiling 5.7 \\
    SVG2 & 138.6 & 2.51$\times$ [2.45, 2.57] & 32.9\% & 106.0 & 32.6 (23.6\%) & k-means 17.8, plan 7.2, permutation 3.3 \\
    SVOO & 152.6 & 2.25$\times$ [2.20, 2.30] & 37.2\% & 116.3 & 36.3 (23.8\%) & plan 27.6 \\
    SVOO, reclustering call & 425.3 & 0.78$\times$ & 37.2\% & 116.9 & 308.5 & co-clustering 272.2 \\
    SVOO, amortized & 159.4 & 2.15$\times$ & -- & -- & 43.2 (27.1\%) & -- \\
    PISA & 129.8 & 2.54$\times$ [2.53, 2.55] & 35.5\% & 122.2 & 7.7 (5.9\%) & pooling 2.7 \\
    SVG-EAR & 122.7 & 2.82$\times$ [2.76, 2.88] & 22.9\% & 71.9 & 50.8 (41.4\%) & k-means 16.9, compensation 13.8, routing 6.5, plan 6.4 \\
    \rowcolor{pqrsblue} \textbf{PQRS} & \textbf{97.8} & \textbf{3.52$\times$} [3.45, 3.59] & 23.2\% & 66.0 & 31.8 (32.5\%) & selection 12.3, k-means 6.5, compensation 4.7, scoring 3.1 \\
    \textbf{PQRS}, exact offsets & 163.3 & 2.14$\times$ [2.09, 2.19] & 23.2\% & 65.9 & 97.4 & exact-offset pass 53.3, task tables 5.5 \\
    \bottomrule
  \end{tabular}
\end{table*}

To time all methods on the same inputs, we record $Q$, $K$ and $V$ along the dense trajectories of three prompts. Because these inputs come from dense trajectories, the methods with adaptive budgets run at slightly higher densities than in \Cref{tab:main}. As \Cref{tab:supp-e11} shows, PQRS routes every query individually, yet its stages outside attend take about as long as those of SVG2, which routes query clusters. The last row runs PQRS with the exact-offset pass instead of the cached LSE.

\subsection{Qualitative Results}
\label{sec:supp-qual}

\begin{figure*}[t]
  \centering
  \includegraphics[width=\linewidth]{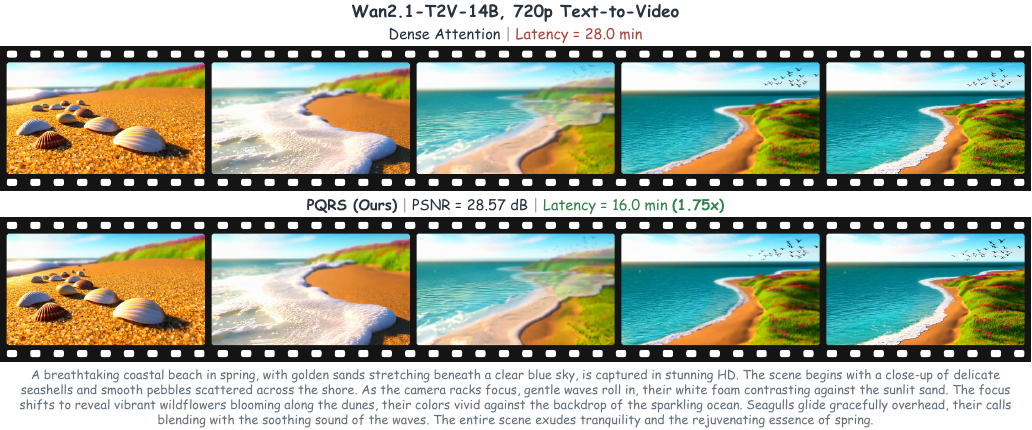}
  \caption{\textbf{Wan2.1-T2V-14B.} Dense attention (top) and PQRS (bottom) with the same prompt and seed.}
  \label{fig:supp-qual-wan14b}
\end{figure*}

\begin{figure*}[t]
  \centering
  \includegraphics[width=\linewidth]{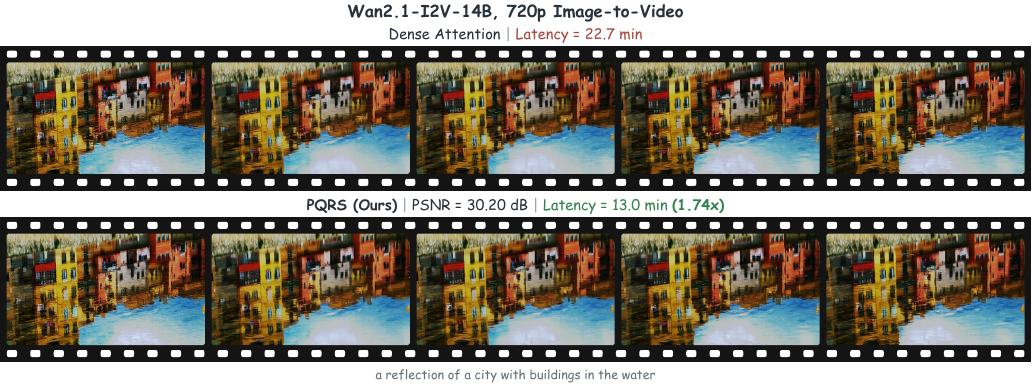}
  \caption{\textbf{Wan2.1-I2V-14B.} Dense attention (top) and PQRS (bottom) with the same input image, prompt and seed.}
  \label{fig:supp-qual-i2v14b}
\end{figure*}

\begin{figure*}[t]
  \centering
  \includegraphics[width=\linewidth]{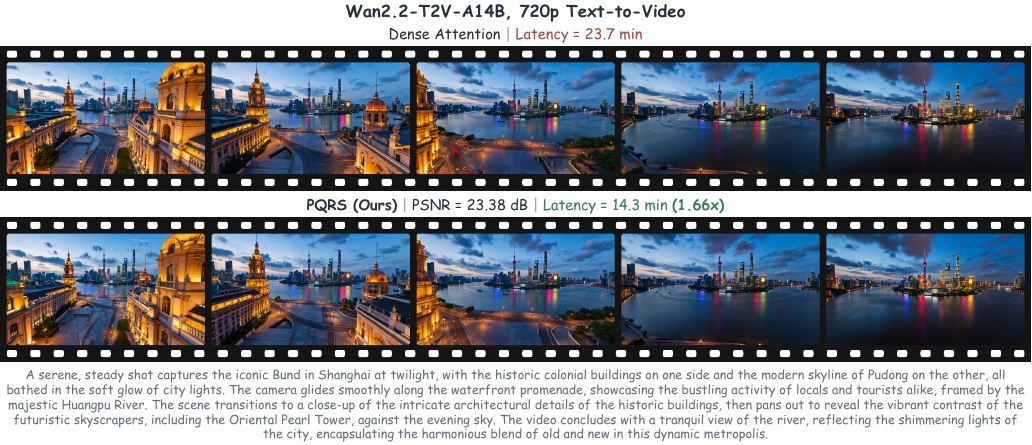}
  \caption{\textbf{Wan2.2-T2V-A14B.} Dense attention (top) and PQRS (bottom) with the same prompt and seed.}
  \label{fig:supp-qual-wan22}
\end{figure*}

\begin{figure*}[t]
  \centering
  \includegraphics[width=\linewidth]{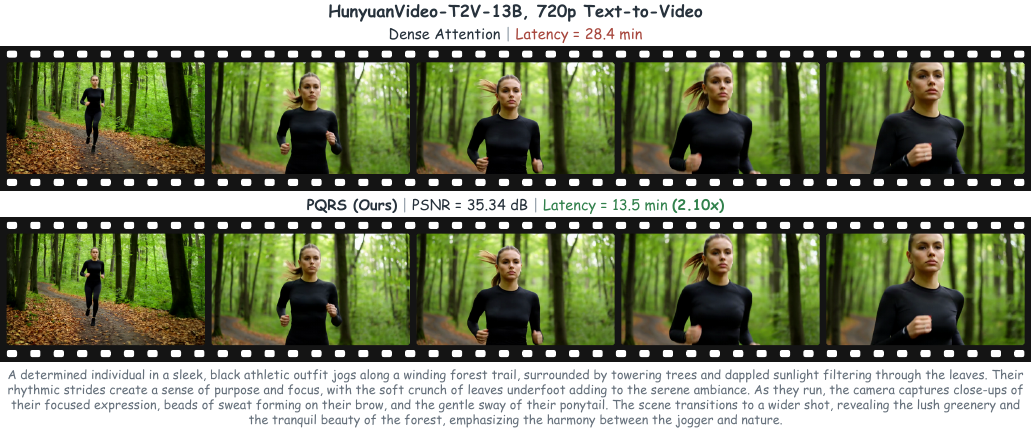}
  \caption{\textbf{HunyuanVideo-T2V-13B.} Dense attention (top) and PQRS (bottom) with the same prompt and seed.}
  \label{fig:supp-qual-hunyuan}
\end{figure*}

\Cref{fig:supp-qual-wan14b,fig:supp-qual-i2v14b,fig:supp-qual-wan22,fig:supp-qual-hunyuan} compare dense attention and PQRS on the four models. For each model we show the prompt on which the PSNR of PQRS is closest to its median over the 50 prompts, excluding the prompts of \Cref{fig:teaser} and two prompts that describe copyrighted characters. Each strip shows five evenly spaced frames, and the latencies are the means of \Cref{tab:main}.

\paragraph{Prompts of \Cref{fig:teaser}.} Wan2.1 (left): ``A charming raccoon, wearing a tiny sailor hat and a striped shirt, strums a miniature guitar while sitting in a small wooden boat. The boat gently rocks on the calm, azure ocean under a clear, sunny sky. The raccoon's nimble fingers pluck the strings with surprising skill, creating a cheerful melody that echoes across the water. Seagulls fly overhead, and the distant horizon is dotted with fluffy white clouds. The raccoon's eyes sparkle with joy as it plays, its bushy tail swaying in time with the music, creating a whimsical and heartwarming scene.'' HunyuanVideo (right): ``A sleek, modern bicycle with a matte black frame and aerodynamic design begins its journey on a smooth, sunlit road. The rider, clad in a fitted, neon green cycling suit and helmet, leans forward, gripping the handlebars tightly. The camera captures the initial slow pedal strokes, the wheels spinning with increasing speed. As the bicycle accelerates, the background blurs, emphasizing the rapid motion. The rider's muscles tense and flex, showcasing the effort and determination. The sunlight glints off the bike's frame and the rider's helmet, creating a dynamic interplay of light and shadow. The sound of the wind rushing past and the rhythmic clicking of the gears enhance the sensation of speed and exhilaration.''

